\pdfoutput=1
\documentclass{article}
\makeatletter\def\input@path{{./}{arxiv/}}\makeatother
\usepackage{iclr2027_preprint,times}

\usepackage{amsmath,amsfonts,bm}

\def\eqref#1{equation~\ref{#1}}

\def\1{\bm{1}}

\def\vp{{\bm{p}}}

\def\vz{{\bm{z}}}

\def\mW{{\bm{W}}}

\DeclareMathAlphabet{\mathsfit}{\encodingdefault}{\sfdefault}{m}{sl}
\SetMathAlphabet{\mathsfit}{bold}{\encodingdefault}{\sfdefault}{bx}{n}

\usepackage{hyperref}
\usepackage{url}
\usepackage{graphicx}
\usepackage{booktabs}
\usepackage{multirow}

\usepackage{xcolor}
\DeclareUnicodeCharacter{25CB}{\scalebox{0.8}{$\bigcirc$}}
\DeclareUnicodeCharacter{00D7}{$\times$}

\title{Supervise What Decides Success:\\ Criterion-Aligned Auxiliary Losses \\ for Latent World-Model Planning}

\author{
Takumi Hara \\
Kyoto University \\
Kyoto, Japan \\
\texttt{thara@easter.kuee.kyoto-u.ac.jp}
\AND
Kanata Suzuki \\
Fujitsu Limited \\
Kanagawa, Japan \\
\texttt{}
}

\iclrfinalcopy

\newcommand{\cmark}{\checkmark}
\newcommand{\xmark}{$\times$}
\begin{document}
\maketitle

\begin{abstract}
Latent world models plan by scoring candidate action sequences with distances in latent space. However, task success is judged by physical quantities, which we call the success-criterion quantities. In all four latent world models we examine, the end-effector position is encoded in the latent state with an error larger than the success criterion allows. Such a latent state cannot separate successful candidates from failing ones. We propose an auxiliary loss that uses success-criterion quantities as training targets, whereas existing latent world models take them only as inputs. During training, a linear head on the encoder and predictor outputs regresses the success-criterion quantities, and the regression error is added to the training loss. The head is discarded after training, so the model, its cost, and its inputs at test time are unchanged. This loss alone improves the success rate on PushT and cube by 3.5\% and 3.4\% (absolute), respectively, and both improvements are statistically significant. A success criterion thus specifies what a world model must retain in its latent state, and we show that it can serve directly as a training target.
\end{abstract}

\section{Introduction}\label{sec:intro}

A world model predicts how actions change the world \citep{hafner2019planet,hafner2020dreamer}. A latent world model encodes the observation image into a low-dimensional latent state with an encoder and makes this prediction in latent space \citep{zhou2025dinowm,assran2025vjepa2,maes2026lewm}. Given a goal image of the scene the robot should reach, the planner selects the action sequence whose predicted latent state is closest to the latent state of the goal image. No pixels are reconstructed. The only quantity that ranks candidate action sequences is this latent distance.

Whether a trial succeeds is decided by the success criterion of each benchmark. A success criterion asks whether the positions of the hand and the objects lie within a tolerance of their values at the goal. Since the latent distance alone ranks the candidates, planning can succeed only if a candidate whose physical quantities used in the criterion, which we call the success-criterion quantities, lie within the tolerance is also the one with the smallest latent distance. Nothing in training asks for this condition. The training loss supervises only the next latent state; the success-criterion quantities appear nowhere in the loss. Indeed, in all four latent world models we examine, the hand position read out from the latent state by linear regression deviates from the true hand position by more than the tolerance of the success criterion. This deviation does not vanish with representation size or pretraining.

Prior work that uses physical quantities, including the success-criterion quantities, in training falls into two groups. DINO-WM \citep{zhou2025dinowm} and V-JEPA 2-AC \citep{assran2025vjepa2} concatenate the proprioceptive state, such as the end-effector position, which we call the hand, to the input. Feeding a quantity as an input does not ask whether the latent state after the predictor retains it at the precision of the success criterion (Section~\ref{sec:pusht}). PhyLatent \citep{zeng2026phylatent} and PSG-JEPA \citep{yan2026psgjepa} use physical quantities as training targets, but the designer chooses which quantities to supervise, and whether the targets correspond to the success criterion is never asked.

\begin{figure}[t]
\centering
\includegraphics[width=\linewidth]{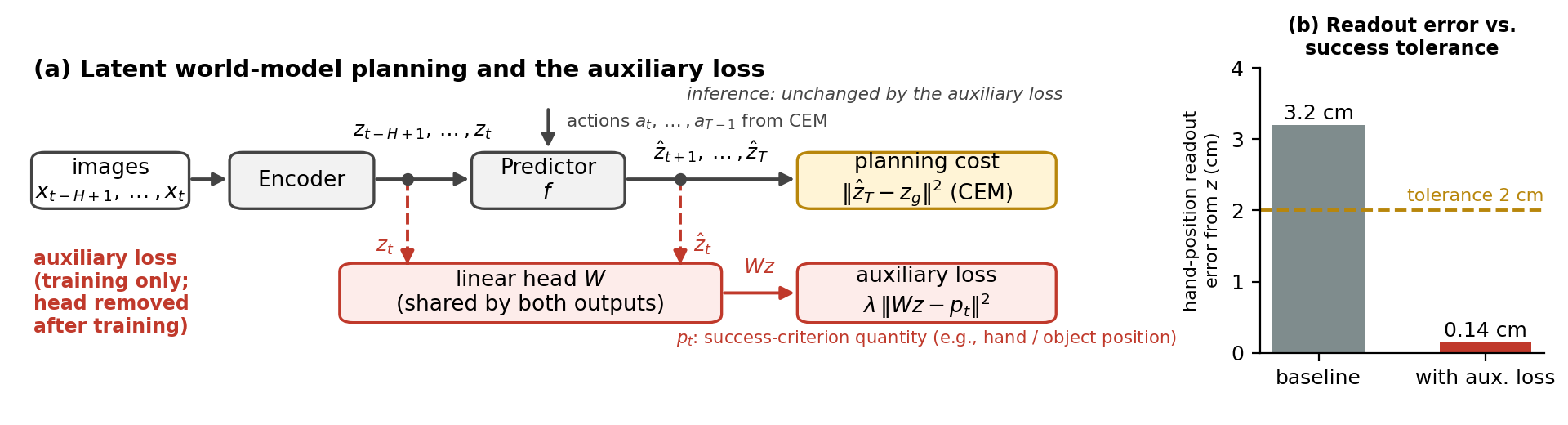}
\caption{\textbf{Planning with a latent world model and the auxiliary loss}. (a) Planning with a latent world model, which is the same with and without the auxiliary loss. Red marks the auxiliary-loss path, which acts only at training time: a shared linear head reads the success-criterion quantities out of the latent states of the encoder and the predictor, and the error is added to the training loss. (b) Readout error of the hand position of LeWM on stack, the median over 5000 images not used for training; the dashed line is the tolerance of the success criterion, 2\,cm. For the model supervised with the hand ($\lambda=1$, Table~\ref{tab:crossreadout}) the error falls inside the tolerance.}
\label{fig:concept}
\end{figure}

We propose an auxiliary loss that uses the success-criterion quantities as training targets. A linear head is applied to both the encoder output and the predictor output, and the error between its readout and the success-criterion quantities recorded in the demonstrations is added to the training loss (Figure~\ref{fig:concept}). The head is discarded after training, so the model, its inputs, and the latent distance used for planning are identical to the baseline at test time, and the computation does not increase. What is supervised is fixed by the success criterion, not chosen by the designer.

The auxiliary loss, however, only drives down the readout error of the success-criterion quantities by one shared linear head from the encoder output and the predictor output. How much a difference in the success-criterion quantities between two candidates shows up as a difference in their latent distances enters the loss nowhere. A corrected readout therefore need not raise the planning success rate. Prior work reports cases where a proxy metric, such as video-prediction quality or linear-probe accuracy, improves while the success rate does not \citep{tian2023vp2,burns2024pvr}, and in our own experiments the frozen V-JEPA 2 shows a corrected readout together with a lower success rate. We therefore evaluate the auxiliary loss by the success rate, and use an ablation over which success-criterion quantities are supervised to determine in which trials the effect appears and which supervision avoids degradation.

Our contributions are threefold.
\begin{itemize}
\item We propose an auxiliary loss that uses the success-criterion quantities as training targets. A linear head is applied to the encoder and predictor outputs only during training, so the model, its inputs, and the latent distance used for planning are unchanged at test time (Section~\ref{sec:method}).
\item On four manipulation tasks, a model supervised with the success-criterion quantities significantly improves the success rate over the baseline, and the improvement replicates on a world model with a frozen pretrained encoder (Section~\ref{sec:pusht}).
\item Through an ablation over which success-criterion quantities are supervised, we show that supervising all quantities in the criterion never significantly degrades the success rate on any task, for both a world model trained from scratch and one with a frozen pretrained encoder (Section~\ref{sec:dissociation}).
\end{itemize}

A success criterion specifies what a world model must retain in its latent state. This work shows that the specification can be used directly as a training target, and delineates how far its effect reaches.

\section{Method}\label{sec:method}

\subsection{Planning in latent space}\label{sec:planning}
A latent world model consists of an encoder and a predictor (Figure~\ref{fig:concept}(a), top). The encoder maps the observation image $x_t$ to a latent state $\vz_t=\mathrm{Enc}(x_t)$. The predictor predicts the next latent state $\hat{\vz}_{t+1}=f(\vz_{t-H+1:t},\,a_{t-H+1:t})$ from the history of latent states and actions, where $H$ is the history length and $a_t$ is the action. For planning, the latent state $\vz_g$ of the goal image serves as the target, and candidate action sequences are evaluated as follows. For each candidate, the predictor is applied sequentially to obtain the terminal latent state $\hat{\vz}_T$. The squared distance $\|\hat{\vz}_T-\vz_g\|^2$ between the terminal and the goal latent states is the cost, and the candidate that minimizes it is found by the cross-entropy method (CEM) \citep{rubinstein1999cross}. CEM ranks candidates by this cost alone, so nothing but the latent distance decides the ranking.

\subsection{Coarse readout of the success-criterion quantities}\label{sec:motivation}
Because candidates are ranked by the latent distance alone, planning can succeed only if a candidate whose success-criterion quantities lie within the tolerance is also the one with the smallest latent distance (Section~\ref{sec:intro}). At the least, this requires that two states whose success-criterion quantities differ by as little as the tolerance be separated by the latent distance near the goal; it does not require the latent distance to agree with physical distance everywhere. The simplest test of this requirement is to read the success-criterion quantities out of the latent state by linear regression. We use ridge regression, a linear regression that penalizes large coefficients, on frames not used for training \citep{alain2017understanding}. On stack, the median error of the hand position read out this way is 3.2\,cm for the baseline LeWM model, and 3.10\,cm, 2.10\,cm, and 4.02\,cm for the image features of three frozen pretrained encoders (DINOv2-S, DINOv2-L, and V-JEPA 2; Section~\ref{sec:setup}), all outside the 2\,cm tolerance. We introduce the auxiliary loss to close this gap. Supervised with the hand, the error of the LeWM model falls to 0.14\,cm (Figure~\ref{fig:concept}(b); Table~\ref{tab:crossreadout} in Appendix~\ref{app:extra}).

\subsection{Auxiliary loss supervised by the success-criterion quantities}\label{sec:method-ground}

The auxiliary loss corrects the coarse readout of Section~\ref{sec:motivation} by restricting the targets to the success-criterion quantities $\vp_t$ and adding only one linear head per quantity and one term to the training loss. For the $k$-th success-criterion quantity $\vp^{(k)}_t$ we place a linear head $\mW_k$ and apply it to both the encoder output $\vz_t$ and the predictor output $\hat{\vz}_t$. The squared error between the head's output and $\vp^{(k)}_t$ is added to the training loss.

\begin{align}
\mathcal{L} \;=\; \mathcal{L}_{\mathrm{wm}}
 \;+\; \textstyle\sum_k \tfrac{\lambda_k}{d_k}\Bigl(\tfrac{1}{N_f}\textstyle\sum_{t=1}^{N_f} \|\mW_k\vz_t-\vp^{(k)}_t\|^2
 \;+\; \tfrac{1}{N_f-1}\textstyle\sum_{t=2}^{N_f} \|\mW_k\hat{\vz}_t-\vp^{(k)}_t\|^2\Bigr).
\end{align}
The second and third terms of Eq.~(1) are the auxiliary loss; they correspond to the paths in Figure~\ref{fig:concept}(a) that are active only during training. The second term is the readout error from the encoder output and the third from the predictor output. $\mathcal{L}_{\mathrm{wm}}$ is the loss the world model has originally; for LeWM it consists of the latent prediction loss and a regularizer on the latent distribution. The auxiliary loss leaves it unchanged. A training sample is a clip of $N_f$ consecutive frames cut from a demonstration. $\hat{\vz}_t$ is the one-step prediction of the latent state at frame $t$ from the frames before it, so a sample yields $N_f-1$ predictions. The second term averages over all $N_f$ frames and the third over the $N_f-1$ predictions. $d_k$ is the number of dimensions of the success-criterion quantity $k$. $k$ indexes the success-criterion quantities and $\lambda_k$ is the weight on quantity $k$; the values are listed in the table in Appendix~\ref{app:impl}.

When there are several success-criterion quantities, the heads $\mW_k$ are independent across quantities. The same $\mW_k$, however, is applied to both the encoder output and the predictor output. The planning cost compares a latent state produced by the predictor with the latent state of the goal image, so unless both outputs hold the success-criterion quantities in a form the same linear head can read, a difference in the quantities does not enter the cost. Sharing the head builds this requirement into the loss.

Rather than using the readout as a planning cost, we discard the linear heads after training. We discard it so that the world model keeps planning from the goal image alone, with a cost over everything the latent state holds; the auxiliary loss only asks that the success-criterion quantities be among what it holds. Inference with the head removed is identical to the baseline: the planning cost, the computation, and the required inputs are unchanged, so no sensor for the success-criterion quantities is needed at test time. The only parameters added during training are those of the linear head.

The targets $\vp_t$ are the physical quantities already recorded in the training data, so training needs no new sensor or measurement. We restrict the targets to the success-criterion quantities because the success criterion specifies what the latent state must retain for planning to succeed, and supervising a quantity outside the criterion does not move the success rate (Section~\ref{sec:pusht}).

\section{Experiments}\label{sec:exp}

\subsection{Experimental setup}\label{sec:setup}\label{sec:method-impl}

\paragraph{Tasks.}We use five tasks (Table~\ref{tab:judges}). PushT \citep{chi2023diffusionpolicy}, cube \citep{park2025ogbench}, and Reacher \citep{tassa2018dmc} are the tasks used in prior work on planning with latent world models such as LeWM and DINO-WM. In PushT a disk pushes a T-shaped block, in cube a block is carried to a target position, and in Reacher the fingertip of a two-joint arm reaches a target. Only PushT among them has both the hand and the object in its success criterion. A task like PushT, where both the hand and the object must lie within tolerance, has a strict criterion and is the right test for the auxiliary loss, but one such task is not enough. We therefore add two tasks from MimicGen \citep{mandlekar2023mimicgen}, which is widely used to evaluate robot learning and synthesizes manipulation demonstrations from a few human demonstrations: stack, which stacks cubes, and coffee, which inserts a pod into a coffee machine.

\paragraph{Unit of evaluation and comparison.}Following prior work, the unit of evaluation is a goal-reaching trial, which we call a trial. As in LeWM, a trial pairs the state at time $t$ of a demonstration in the training data, as the start, with the state the same demonstration reached 25 steps later, as the goal; the planner receives the image of the goal state. Because the goal is a state reached within 25 steps, a failure cannot be attributed to an unreachable goal. Tasks in which reachability decides success, such as mazes or tasks with walls, are outside our scope (Section~\ref{sec:limitations}). We draw 2000 trials at random from the demonstrations and give the same set to the baseline model, trained without the auxiliary loss, and to the model trained with it, which we call the auxiliary-loss model. The choice of trials therefore favors neither model, and differences in difficulty across trials cancel in the comparison. The pair of outcomes of the two models on the same trial is a trial pair, and all comparisons below are made in this unit. A trial may execute up to 50 action steps and succeeds if the success criterion described below (Table~\ref{tab:judges}) holds at any step within this budget.

\paragraph{Success criteria.}The success criteria of all five tasks take the same form: the success-criterion quantities must lie within a tolerance of their goal values (Table~\ref{tab:judges}). For PushT, cube, and Reacher we use the criteria of LeWM's evaluation unchanged. For stack and coffee, MimicGen provides no criterion for the goal-reaching evaluation that prior work has used for PushT, cube, and Reacher, which measures whether a goal state 25 steps ahead is reached, so we defined one of the same form. The object tolerance follows cube, 4\,cm. With a hand tolerance of 4\,cm, many trials were already judged a success at their start. To avoid this, we set the hand tolerance to 2\,cm. At 2\,cm, both such trials and false matches with states from other episodes are rare (Appendix~\ref{app:metrics}). Both tolerances were fixed before the experiments with the auxiliary loss.

\begin{table}[t]
\centering\footnotesize
\caption{Form of the success criteria. The top three rows are the criteria of the evaluation used by LeWM; the bottom two rows are criteria we defined and fixed before the auxiliary-loss experiments. agent is the disk moved by the policy, block the pushed object, obj the stacked object, hand the end effector, $q$ the joint angles of Reacher, pod the coffee pod, goal the value in the respective goal state, and $\Delta\theta$ the difference of the block angle from the goal.}
\label{tab:judges}
\begin{tabular}{lll}
\toprule
Task & Success criterion & Origin \\
\midrule
PushT & $\sqrt{\|\text{agent}-\text{goal}\|^{2}+\|\text{block}-\text{goal}\|^{2}}<20 \,\land\, |\Delta\theta|<20^{\circ}$ & Prior evaluation \\
cube & $\|\text{block}-\text{goal}\|<4\,\mathrm{cm}$ & Prior evaluation \\
Reacher & $|q-q_{\text{goal}}|<0.05\,\mathrm{rad}$ (all joints) & Prior evaluation \\
stack & $\|\text{obj}-\text{goal}\|<4\,\mathrm{cm} \,\land\, \|\text{hand}-\text{goal}\|<2\,\mathrm{cm}$ & Ours \\
coffee & $\|\text{pod}-\text{goal}\|<4\,\mathrm{cm} \,\land\, \|\text{hand}-\text{goal}\|<2\,\mathrm{cm}$ & Ours \\
\bottomrule
\end{tabular}
\end{table}

\paragraph{Model and training.}We evaluate the auxiliary loss on LeWM \citep{maes2026lewm}, a JEPA-type \citep{lecun2022path,assran2023ijepa} world model trained from scratch without pretraining, which predicts latent states rather than images. The training data are successful demonstrations. For PushT, cube, and Reacher we use the datasets distributed by prior work unchanged; for stack and coffee we generated demonstrations with MimicGen. The numbers of demonstrations are listed in Appendix~\ref{app:impl}. The baseline model and the auxiliary-loss model are trained on the same demonstrations for the same 10 epochs, and the only difference between them is the auxiliary loss. For each condition we train one model with one seed. A training sample has $N_f=4$ frames five steps apart, so one application of the predictor predicts the latent state five steps ahead. At planning time, the encoder is applied only to the current image and the goal image, and the predictor is applied five times in succession to its own outputs to cover the 25 steps to the goal. The planned 25 steps are executed before the plan is recomputed from a new image (Section~\ref{sec:planning}). The model architecture, the latent dimension, the image resolution, the number of CEM candidates and iterations, and the auxiliary-loss weights are given in Appendix~\ref{app:impl}.

\paragraph{Frozen-encoder system.}To test whether the auxiliary loss works beyond LeWM, we also use a system we implemented after DINO-WM \citep{zhou2025dinowm}. It freezes a pretrained DINOv2 \citep{oquab2023dinov2} encoder, ViT-S or ViT-L, trains only the predictor, and concatenates the hand position to the predictor's input; we call it the DINO-WM type system. Success rates are never compared across systems; every comparison is between trial pairs within one system.

\paragraph{Statistical test.}The effect of the auxiliary loss is judged by a two-sided paired McNemar test \citep{mcnemar1947test} on the same trial pairs. Effect sizes are absolute differences in success rate, written in \%.

\subsection{Success rate by supervised quantity}\label{sec:pusht}

\paragraph{Improvement on every manipulation task.}Table~\ref{tab:official} shows how the success rate changes when the supervised success-criterion quantities are varied, for the five tasks. In each of the four manipulation tasks, some model supervised with success-criterion quantities improves the success rate significantly. Which quantities must be supervised differs by task. In stack, cube, and PushT the improving model is the one supervised with all quantities in the criterion; in cube the criterion contains only the block position, so this is the object-only model. In coffee, the object-only model improves while the model supervised with both does not. The improvements are 3--5\% absolute, on baselines from 30\% in stack to over 90\% in PushT. Why the required supervision differs by task is examined per trial stratum in Section~\ref{sec:dissociation}.

\paragraph{Replication on a frozen-encoder system.}The improvement is not specific to LeWM, whose encoder is trained together with the predictor. In the DINO-WM type system, supervising the hand position on stack improves the success rate significantly with both encoder sizes, ViT-S and ViT-L (Table~\ref{tab:official}). The baseline of this system receives the hand position as an input to the predictor, yet the hand position read out from its latent state deviates by more than the tolerance of the success criterion (Section~\ref{sec:motivation}). Feeding a success-criterion quantity as an input therefore does not make the latent state retain it at the precision the criterion requires; the success rate rises only when the quantity is used as a training target. Replication on the frozen-encoder system is limited to stack; measurements on PushT and on a frozen V-JEPA 2 are given in Appendix~\ref{app:vjepa}.

\paragraph{Absence of degradation.}The models supervised with all quantities in the criterion never degrade the success rate significantly. In stack, cube, and PushT they improve it, and on coffee the difference is $-0.60$\%, not a significant degradation (Table~\ref{tab:official}). The five stack models supervised with both quantities under different weights and training seeds do not degrade it significantly either (Appendix~\ref{app:extra}). The same holds in the frozen-encoder system, where supervising both quantities on stack improves the success rate by 4.20\% ($n=1000$, Table~\ref{tab:official}).

\paragraph{Supervision that leaves the success rate unchanged.}Supervising a quantity outside the success criterion, in contrast, does not move the success rate. The criterion of cube contains only the block position and that of Reacher only the joint angles, so the hand in cube and the fingertip in Reacher lie outside the criterion, and the models supervised with them show no significant difference (Table~\ref{tab:official}). The joint angles of Reacher are in the criterion, but the recorded angles are not wrapped, so the same pose can carry values $2\pi$ apart and the angles cannot be determined uniquely from the image; supervising them does not move the success rate either (Appendix~\ref{app:impl}). Among quantities in the criterion, four cases leave the overall success rate unchanged: the object in PushT, the hand in coffee, and the hand and the object in stack (Table~\ref{tab:official}). The reasons for these four cases are given in Section~\ref{sec:dissociation}.

\begin{table}[t]
\centering
\caption{\textbf{Change in success rate}. Success rate of the baseline model and the difference of the auxiliary-loss model from it (absolute, \%). LeWM uses identical pairs of 2000 trials; the DINO-WM type identical pairs of 1000 trials. Criteria as in Table~\ref{tab:judges}; test: paired two-sided McNemar; $^{*}$ marks $p<0.05$. ``Precondition'' is whether the supervised quantity is in the success criterion and visible in the observed image; the three arms below the double rule do not satisfy it (the hand in cube and the fingertip in Reacher are not in the criterion, and the joint angles of Reacher are not visible in the image). One training seed per model; auxiliary-loss weights in Appendix~\ref{app:impl}.}
\label{tab:official}
\footnotesize
\begin{tabular}{llcc}
\toprule
Task (baseline success rate) & Supervised quantity & Precondition & Diff. [\%] \\
\midrule
stack (30.1\%) & hand only & ○ & $-1.30$ \\
 & object only & ○ & $-0.40$ \\
 & both & ○ & $+5.05^{*}$ \\
\midrule
coffee (34.0\%) & hand only & ○ & $+0.00$ \\
 & object only & ○ & $+3.90^{*}$ \\
 & both & ○ & $-0.60$ \\
\midrule
cube (68.5\%) & object only & ○ & $+3.40^{*}$ \\
 & both & ○ & $+3.25^{*}$ \\
\midrule
PushT (91.5\%) & hand only & ○ & $+2.00^{*}$ \\
 & object only & ○ & $+0.20$ \\
 & both & ○ & $+3.50^{*}$ \\
\midrule
DINO-WM type (DINOv2-S), stack (32.0\%) & hand only & ○ & $+7.10^{*}$ \\
 & object only & ○ & $-7.70^{*}$ \\
 & both & ○ & $+4.20^{*}$ \\
DINO-WM type (DINOv2-L), stack (34.7\%) & hand only & ○ & $+3.60^{*}$ \\
\midrule\midrule
cube (68.5\%) & hand only & × & $+0.25$ \\
Reacher (85.0\%) & fingertip & × & $-1.25$ \\
 & joint angles & × & $-1.70$ \\
\bottomrule
\end{tabular}
\end{table}

\subsection{Effect per trial stratum}\label{sec:dissociation}

\begin{figure}[t]
\centering
\includegraphics[width=0.9\linewidth]{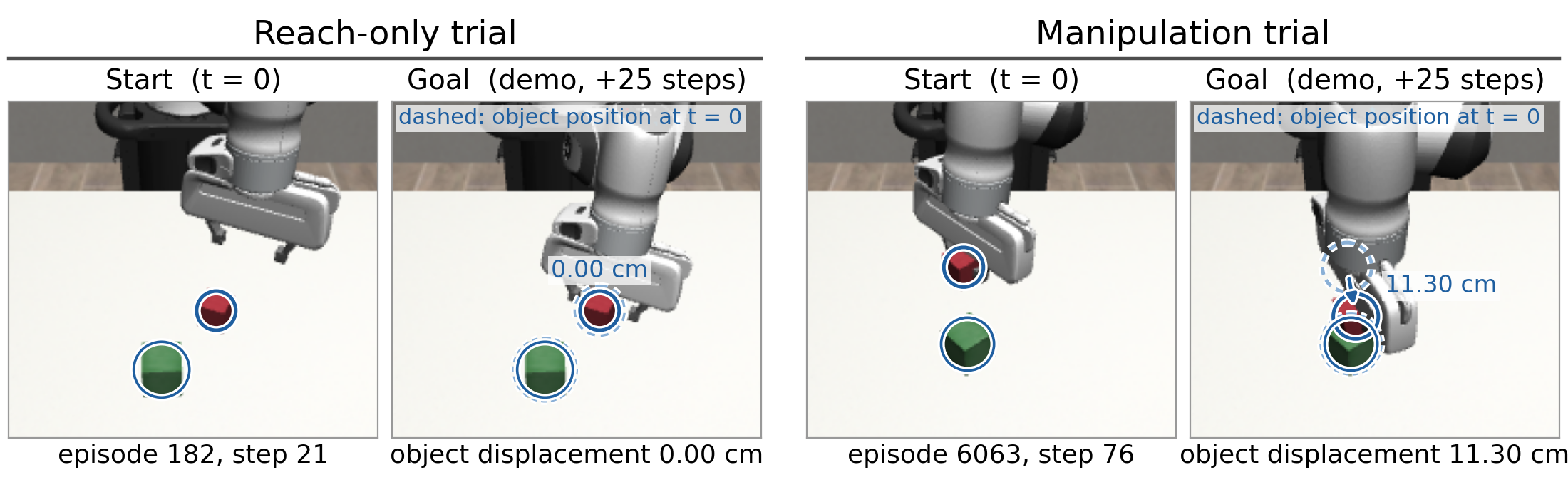}
\caption{\textbf{Examples of the trial strata (stack)}. The start frame and the goal frame of the demonstration for a reach-only trial and a manipulation trial. The strata are those of Table~\ref{tab:dissociation}; a trial with an object displacement of at least 4\,cm is a manipulation trial. Solid circles mark the cube positions, dashed circles the positions at the start, and the number by the arrow is the displacement from start to goal. In the reach-only trial the dashed circle coincides with the solid one. Examples for all five tasks are in Figure~\ref{fig:strata-all} of Appendix~\ref{app:metrics}.}
\label{fig:strata}
\end{figure}

\paragraph{Trial strata.}The overall success rate hides the effect of each supervised quantity. Some trials succeed by moving only the hand, while others also require carrying the object. We therefore split the trials into these two kinds and examine the effect of each supervised quantity separately. The goal of a trial is the state the demonstration reached 25 steps after the start (Section~\ref{sec:setup}). If the object's position at the start lies within the object tolerance of the criterion (Table~\ref{tab:judges}) of its goal position, the object already satisfies its condition, and the trial succeeds once the hand reaches its goal; we call these reach-only trials. If the difference is at or above the tolerance, the object must be carried to its goal; we call these manipulation trials. The split depends only on the start and goal states and not on the model evaluated. Figure~\ref{fig:strata} shows examples from stack. In PushT the agent plays the role of the hand. In cube the criterion contains only the block position, so reach-only trials are already successes at the start and only manipulation trials are compared.

\paragraph{Complementary degradation on stack.}Table~\ref{tab:dissociation} gives the success rates per stratum. In stack with LeWM, which stratum improves depends on the supervised quantity. Supervising the hand alone improves reach-only trials and degrades manipulation trials; supervising the object alone does the opposite, degrading reach-only trials and improving manipulation trials. Reach-only trials test only the hand condition, and manipulation trials additionally test the object condition, so on stack with LeWM the auxiliary loss helps in the trials that test the supervised quantity and hurts in the others. We call the degradation on the unsupervised side complementary degradation. In the trials that the baseline also fails, the hand-only model brings the hand about 2\,cm closer to its goal than the baseline in manipulation trials but leaves the cube about 10\,cm short, and the object-only model ends reach-only trials with the hand still outside the 2\,cm tolerance (Appendix~\ref{app:extra}).

\paragraph{Cancellation in the overall success rate.}In stack with LeWM, the hand-only and object-only models appear ineffective when only the overall success rate is examined without the strata (Table~\ref{tab:dissociation}), because the significant improvement in one stratum and the significant degradation in the other cancel when added. In stack with the frozen DINOv2-S, no cancellation occurs: supervising the hand alone improves both strata, and supervising the object alone degrades reach-only trials so strongly that the overall success rate falls too (Table~\ref{tab:dissociation}). In this system the baseline succeeds in only 5.7\% of manipulation trials, so changes in reach-only trials almost entirely determine the overall success rate.

\paragraph{Absence of degradation per stratum.}If the degradation occurs on the unsupervised side, supervising all quantities in the criterion should remove it. Section~\ref{sec:pusht} showed that the overall success rate never degrades significantly; the same holds per stratum. The stack model supervised with both quantities improves reach-only trials and does not degrade manipulation trials (Table~\ref{tab:dissociation}). In coffee the difference in manipulation trials remains negative with both supervised, but it is not a significant degradation. Across cube and PushT as well, none of the eight models supervised with all quantities in the criterion degrades either stratum significantly (Table~\ref{tab:dissociation}, Appendix~\ref{app:extra}). The same holds in the frozen DINOv2-S system, where supervising both on stack degrades neither stratum significantly (Table~\ref{tab:dissociation}).

\paragraph{Task dependence of complementary degradation.}Complementary degradation is an observation on stack and does not hold in every task. Supervising the hand alone degrades manipulation trials on coffee as well (Table~\ref{tab:dissociation}), but supervising the object alone degrades reach-only trials only on stack. On coffee, supervising the object alone even improves reach-only trials: on stack the effect appeared only in the trials that test the supervised quantity, whereas on coffee supervising the object helps trials that test only the hand. In PushT neither the hand-only nor the object-only model degrades either stratum significantly, and on cube manipulation trials do not degrade (Table~\ref{tab:dissociation}). Two of the differences across tasks are explained by task conditions. The criterion of PushT folds the agent and block errors into one distance dominated by the agent error, so supervising the object alone does not move the success rate (Appendix~\ref{app:extra}). The criterion of cube has no hand term, so no stratum tests the hand. On coffee, supervising the hand alone does not move the success rate because coffee has few failures of the kind that hand supervision converts. On stack, most failures in reach-only trials stop a few centimeters short of the goal, and the hand-only models turn such failures into successes. On coffee, only 12\% of the reach-only failures stop short; 44\% never come within 4\,cm of the goal, and 44\% come closest at the end of the first plan and then move away after replanning. Supervising the hand does not change the shares of these two types (Appendix~\ref{app:extra}).

\paragraph{Readout accuracy versus success rate.}A more accurate readout does not by itself raise the success rate. The model supervised only with the object reads out the hand position from the terminal latent state that enters the planning cost more accurately than the baseline (Appendix~\ref{app:extra}). The hand information is therefore in the latent state, and it survives the predictor. Nevertheless, the success rate of this model on reach-only trials falls.

\begin{table}[t]
\centering\footnotesize
\caption{\textbf{Effect per trial stratum (four tasks)}. Absolute difference in success rate from the baseline model [\%]. Trial pairs, criteria, test, seeds, and weights as in Table~\ref{tab:official}. Baseline rows give the baseline success rate and the $n$ of each stratum. A trial is a manipulation trial if the object displacement between start and goal is at least the object tolerance of the criterion (stack and coffee 4\,cm, PushT 20\,units), and a reach-only trial otherwise. Reach-only trials of cube satisfy the criterion at the start because it has no hand term, so they are not compared (---). The 8 stack models with varied weights and seeds are in Appendix~\ref{app:extra}.}
\label{tab:dissociation}
\begin{tabular}{llccc}
\toprule
Task & Supervised quantity & Overall & Reach-only & Manipulation \\
\midrule
stack & baseline & 30.1\% & 44.8\% ($n=980$) & 16.1\% ($n=1020$) \\
 & hand only & $-1.30$ & $+5.92^{*}$ & $-8.24^{*}$ \\
 & object only & $-0.40$ & $-7.86^{*}$ & $+6.76^{*}$ \\
 & both & $+5.05^{*}$ & $+9.08^{*}$ & $+1.18$ \\
\midrule
coffee & baseline & 34.0\% & 41.7\% ($n=1296$) & 19.9\% ($n=704$) \\
 & hand only & $+0.00$ & $+2.01$ & $-3.69^{*}$ \\
 & object only & $+3.90^{*}$ & $+5.25^{*}$ & $+1.42$ \\
 & both & $-0.60$ & $+0.62$ & $-2.84$ \\
\midrule
cube & baseline & 68.5\% & —\,($n=745$) & 49.7\% ($n=1255$) \\
 & hand only & $+0.25$ & — & $+0.40$ \\
 & object only & $+3.40^{*}$ & — & $+5.42^{*}$ \\
 & both & $+3.25^{*}$ & — & $+5.18^{*}$ \\
\midrule
PushT & baseline & 91.5\% & 92.3\% ($n=862$) & 90.8\% ($n=1138$) \\
 & hand only & $+2.00^{*}$ & $+1.97^{*}$ & $+2.02^{*}$ \\
 & object only & $+0.20$ & $-1.28$ & $+1.32$ \\
 & both & $+3.50^{*}$ & $+4.18^{*}$ & $+2.99^{*}$ \\
\midrule
DINO-WM type & baseline & 32.0\% & 61.2\% ($n=474$) & 5.7\% ($n=526$) \\
 (DINOv2-S), stack & hand only & $+7.10^{*}$ & $+9.28^{*}$ & $+5.13^{*}$ \\
 & object only & $-7.70^{*}$ & $-19.83^{*}$ & $+3.23$ \\
 & both & $+4.20^{*}$ & $+2.74$ & $+5.51^{*}$ \\
\bottomrule
\end{tabular}
\end{table}

\section{Related work}\label{sec:related}

\paragraph{Planning with latent world models.}World models that plan in latent space map observations to latent states, predict ahead with a predictor, and rank action sequences by the distance between latent states \citep{hafner2019planet,hafner2020dreamer,hafner2025dreamerv3,micheli2023iris,hansen2024tdmpc2,zhou2025dinowm,assran2025vjepa2,maes2026lewm}. In none of them is a success-criterion quantity a training target. Some give it as an input: DINO-WM and V-JEPA 2-AC concatenate the proprioceptive state to the latent state. Some recover it as part of observation reconstruction: PlaNet and Dreamer reconstruct pixels. In LeWM it does not enter the model at all. A further line provides information available only at training time, to the critic in asymmetric actor-critic \citep{pinto2018asymmetric}, to the teacher policy in Learning by Cheating \citep{chen2019lbc}, and to the encoder of environment factors in RMA \citep{kumar2021rma}. In all of these, the designer chooses what to provide. None has a loss term that asks whether the success-criterion quantities can be read from the latent state at the precision of the criterion. We add that term, and the success criterion, not the designer, decides what is supervised. LeWM rejects adding auxiliary targets as a means of preventing representational collapse. Our auxiliary loss is not for preventing collapse but for making the success-criterion quantities readable from the latent state. World-model benchmarks \citep{jiang2026robowmbench,shang2026worldarena} aim to rank models by capability, not to measure the size of an intervention's effect. Work that detects failures at test time from the latent state also exists \citep{zhang2026foresight}, whereas our intervention acts at training time.

\paragraph{Shaping representations with auxiliary losses.}Studies that shape representations with auxiliary losses differ in what they supervise. The closest, RC-aux \citep{li2026rcaux}, supervises reachability, a relational quantity. Like ours, it places heads on the outputs of both the encoder and the predictor, but it stops the gradient on the predictor side, whereas we do not. RC-aux reports that the readout of physical quantities does not change with its intervention, and we do not supervise reachability, so the two interventions act on different things. Concurrently, PhyLatent \citep{zeng2026phylatent} regresses the full simulator state as one term of a combined loss on both outputs, and PSG-JEPA \citep{yan2026psgjepa} supervises only the proprioceptive state, only at the encoder output. Neither contrasts quantities inside the criterion with quantities outside it, nor measures the effect per trial stratum. More broadly, representations have been shaped by self-prediction, augmentation invariance, reward and pixels, positions, regression of physical quantities, and auxiliary tasks \citep{schwarzer2021spr,laskin2020curl,seo2022mwm,ferraro2024lcp,levine2016end,jaderberg2017unreal}, and DBC \citep{zhang2021dbc} defines the latent distance so that states with similar rewards and transitions are close. These works aim at the quality of the representation handed to a policy or a learner; whether the latent distance used as the planning cost reflects the success criterion is neither their goal nor their evaluation. Table~\ref{tab:positioning} in Appendix~\ref{app:positioning} compares the three closest works item by item.

\paragraph{Learning the distance used as a cost.}Some prior work learns the distance used as a planning cost itself from data. VIP and LIV \citep{ma2023vip,ma2023liv} learn an image embedding in which frames close in time within a video are close, and use the embedding distance to the goal image as a reward or a cost. Quasimetric RL \citep{wang2023qrl} learns, as the distance, the cost of getting from one state to another. TRM \citep{li2026trm} re-learns the terminal cost of a fixed world model from the outcomes of executed trajectories. In none of these is the distance learned from the success criterion, so there is no guarantee that differences in success-criterion quantities are reflected in the distance. We leave the cost unchanged and instead restrict the training targets of the latent state to the success-criterion quantities, which lets us ask directly whether those differences are reflected in the distance.

\section{Conclusion and limitations}\label{sec:limitations}

\paragraph{Conclusion.}We proposed an auxiliary loss that uses only the success-criterion quantities as training targets, and measured its effect on five tasks with LeWM, a world model trained from scratch. In all four tasks whose success-criterion quantities can serve as training targets, a model supervised with them improved the success rate. Supervising all quantities in the criterion never significantly degraded the success rate, overall or per trial stratum, whereas supervising a quantity outside the criterion left it unchanged. Supervising only some of the quantities helped some trials and, on stack, hurt the others. A corrected readout alone did not raise the success rate. The improvement further replicated on stack with a world model whose pretrained encoder is frozen, where supervising all quantities in the criterion again caused no significant degradation. A success criterion specifies what a world model must retain in its latent state, and this work shows that the specification can be used directly as a training target. In practice, the results give a simple rule: when a task comes with a success criterion, supervise every quantity in it. The rule adds nothing to the model at test time, and supervising all the quantities never significantly degraded the success rate, so it can be applied without first finding out which quantity matters.

\paragraph{Limitations.}Three limitations remain. First, each condition has a single training seed, so the comparisons rest on paired tests over identical trial pairs (Appendix~\ref{app:metrics}) \citep{colas2018seeds,agarwal2021rliable}. The improvements of the five all-quantity stack models range from $+1.35$\% to $+11.35$\% (Appendix~\ref{app:extra}). Second, the auxiliary-loss weight was swept only for the hand in stack (Appendix~\ref{app:extra}). Third, replication on a frozen-encoder world model is limited to stack, and it failed on a frozen V-JEPA 2 (Appendix~\ref{app:vjepa}).

\paragraph{Future work.}Four questions remain. First, for criteria defined by predicates such as contact or grasp rather than by quantities, what to supervise is undetermined. Second, on real robots, success-criterion quantities that kinematics cannot provide, such as object positions, must be obtained by motion capture or estimators. Third, the route by which the auxiliary loss raises the success rate is not identified; the loss improves the ranking of candidate action sequences, but the size of that improvement does not track the size of the success-rate gain (Appendix~\ref{app:extra}). Fourth, the degradation caused by supervising only some of the quantities is not explained by a worse readout of the unsupervised quantities (Section~\ref{sec:dissociation}), and its mechanism is unresolved.

\subsection*{AI use statement}
In this work, we used large language models for tasks with required disclosure: assisting with translation from Japanese to English, implementing analysis code, providing feedback on the design of experiments, and supporting data analysis. We did not use them to generate data, to develop theoretical models, or to formulate or prove mathematical claims. Additionally, we used them for tasks with recommended disclosure: drafting and editing the text, creating figures, and surveying related work. We have reviewed all AI-assisted work: the analysis code was verified against the recorded data, and every claim was checked against the measurements reported in this paper. The interpretation of the results and all decisions on the experiments and the claims were made by the authors, who take responsibility for the final content of this work.

\subsubsection*{Acknowledgments}
This work was supported by JST PRESTO, Japan, Grant Number JPMJPR24T4.

\bibliography{iclr2027_conference}
\bibliographystyle{iclr2027_conference}

\appendix

\section{Implementation details}\label{app:impl}
\paragraph{Model and planning.}Of the settings of the model and the planner in Section~\ref{sec:method-impl}, we record here the values not given in the main text. The encoder of LeWM is a ViT-Tiny \citep{dosovitskiy2021vit} with 5.5M parameters and the predictor a causal Transformer with 10.8M parameters. The latent state has 192 dimensions and the image resolution is 224$\times$224. One application of the predictor takes the actions of five consecutive steps together and predicts the latent state five steps ahead. Planning searches over action sequences of five predictor applications, that is, 25 steps, and executes them before replanning. CEM uses 300 candidates and 30 iterations in LeWM. At each iteration it refits the mean and standard deviation of the action distribution to the top 30 candidates, and after the last iteration the mean of the top 30 candidates is executed. Because the forward pass of the DINO-WM type and the frozen V-JEPA 2 is expensive, we reduce the budget to 150 candidates and 15 iterations on stack, and keep 300 candidates and 30 iterations on PushT as in LeWM. The DINO-WM type feeds the predictor with the output of a frozen DINOv2 encoder (384 dimensions for ViT-S, 1024 for ViT-L) concatenated with the proprioceptive state (19 dimensions), and trains only the predictor for 10 epochs with the same settings as LeWM. The frozen V-JEPA 2 system is the same except that the encoder is replaced by V-JEPA 2 (ViT-L, 1024 dimensions), pretrained on video, and a still image is duplicated into the two frames it expects.
\paragraph{Linear head and squared error.}We use a linear head because the planning cost is the Euclidean distance in latent space, so a difference in a success-criterion quantity must be readable from the latent state linearly to appear in that distance (Section~\ref{sec:motivation}). The head is a linear map from the latent state to the quantity with a bias term, shared between the encoder output and the predictor output. The loss is the squared error, and the tolerance of the criterion does not enter the loss. The targets are standardized over the training data so that the value of $\lambda$ does not depend on the units of the quantity.

\paragraph{Auxiliary-loss weights and added parameters.}Table~\ref{tab:lambda} lists the auxiliary-loss weights $\lambda$ and the number of parameters added by the linear heads, which is the weights plus the bias of each head. The input dimension of the head is 192, the latent state, in LeWM. In the DINO-WM type it is 394 or 1034, the image feature of 384 (ViT-S) or 1024 (ViT-L) concatenated with a 10-dimensional embedding of the proprioceptive state; in the frozen V-JEPA 2 it is 1034. All DINO-WM-type and frozen V-JEPA 2 models use $\lambda=1$. For the hand in stack we tried three weights, 1, 0.1, and 0.03, and compared them on the same 2000 trials used for evaluation. Table~\ref{tab:official} shows the hand weight $\lambda=1$, the same as in the model supervised with both quantities. For the other tasks and systems we did not sweep the weight and used either the default values, 1 for the hand and 0.1 for the object, or a value tried on stack.

\begin{table}[h]
\centering
\caption{\textbf{Auxiliary-loss weights and added parameters}. For the auxiliary-loss models in Tables~\ref{tab:official} and \ref{tab:dissociation}, the weights $\lambda$ for the hand and the object and the number of parameters added by the linear heads. --- means the head is not used. The $\lambda$ of the 8 stack models with varied weights and seeds are given in Figure~\ref{fig:stack12}.}
\label{tab:lambda}
\small
\begin{tabular}{lllccr}
\toprule
System & Task & Supervised quantity & $\lambda$ hand & $\lambda$ object & Added parameters \\
\midrule
LeWM & PushT & hand only & 1 & --- & 386 \\
 & & object only & --- & 0.1 & 579 \\
 & & both & 1 & 0.1 & 965 \\
LeWM & cube & hand only & 0.03 & --- & 579 \\
 & & object only & --- & 0.1 & 579 \\
 & & both & 1 & 0.1 & 1,158 \\
LeWM & stack & hand only & 1 & --- & 579 \\
 & & object only & --- & 0.1 & 1,158 \\
 & & both & 1 & 0.1 & 1,737 \\
LeWM & coffee & hand only & 0.03 & --- & 579 \\
 & & object only & --- & 0.1 & 1,158 \\
 & & both & 0.03 & 0.1 & 1,737 \\
LeWM & Reacher & joint angles & 1 & --- & 386 \\
 & & fingertip & 1 & --- & 386 \\
\midrule
DINO-WM type (DINOv2-S) & stack & hand only & 1 & --- & 1,185 \\
 & & object only & --- & 1 & 2,370 \\
 & & both & 1 & 1 & 3,555 \\
DINO-WM type (DINOv2-S) & PushT & hand only & 1 & --- & 790 \\
DINO-WM type (DINOv2-L) & stack & hand only & 1 & --- & 3,105 \\
Frozen V-JEPA 2 & stack & hand only & 1 & --- & 3,105 \\
Frozen V-JEPA 2 & PushT & hand only & 1 & --- & 2,070 \\
\bottomrule
\end{tabular}
\end{table}

\paragraph{Tasks and training data.}Table~\ref{tab:tasks} lists the environment and the training data of each task. The observation is a 224$\times$224 RGB image in all tasks. PushT pushes a T-shaped object with a disk; cube is cube-single of OGBench and carries a block to a target position; Reacher is the two-joint arm of the DeepMind Control Suite \citep{tassa2018dmc} and brings the fingertip to a target. For these three tasks we use the datasets distributed by LeWM unchanged; the PushT data originate from Diffusion Policy, the cube data from OGBench, and the Reacher data were collected by the authors of LeWM. stack and coffee are robosuite \citep{zhu2020robosuite} environments with demonstrations synthesized by MimicGen from a few human demonstrations. Trials were made by taking, in each demonstration, the time steps whose state 25 steps later lies within the same demonstration as valid start points, and drawing 2000 of them at random without replacement. The set of trials is identical for the baseline and the auxiliary-loss models. The joint angles of Reacher are recorded as accumulated rotation, not folded into the range from $-\pi$ to $\pi$. The same pose therefore has target values $2\pi$ apart, and 13.6\% of the frames lie beyond $\pi$. A linear head cannot read such a target from the image: the readout error of the trained head is 0.53\,rad in median, ten times the tolerance of 0.05\,rad (Section~\ref{sec:pusht}).

\begin{table}[h]
\centering
\caption{\textbf{Tasks and training data}. Total frames is the number of images over all time steps of all demonstrations in the training data. Numbers of demonstrations and total frames are measured on the training data; action dimensions follow the environment definitions.}
\label{tab:tasks}
\small
\begin{tabular}{llrrr}
\toprule
Task & Source of demonstrations & Demos & Total frames & Action dim. \\
\midrule
PushT & Diffusion Policy (redistributed by LeWM) & 18,685 & 2,336,736 & 2 \\
cube & OGBench (redistributed by LeWM) & 10,000 & 2,010,000 & 5 \\
Reacher & collected by the LeWM authors & 10,000 & 2,010,000 & 2 \\
stack & synthesized by MimicGen & 20,000 & 2,196,344 & 7 \\
coffee & synthesized by MimicGen & 20,000 & 4,371,670 & 7 \\
\bottomrule
\end{tabular}
\end{table}

\section{Frozen-encoder systems: scope of replication and an exception}\label{app:vjepa}
\paragraph{Scope of replication.}Replication on the frozen-encoder systems is limited to stack. On PushT, supervising the hand changes the success rate by $+1.45$\% ($p=0.0503$) on the DINO-WM type with DINOv2-S, whose baseline is 87.95\%, and by $-1.40$\% ($p=0.22$) on the frozen V-JEPA 2, whose baseline is 45.05\%; neither is significant at $n=2000$.

\paragraph{The exception of the frozen V-JEPA 2.}The frozen V-JEPA 2 is the exception to the replication: its success rate falls even though the readout improves. When the hand is supervised on stack, the readout error of the hand position from the predicted latent state falls from 1.18\,cm to 0.58\,cm, but the success rate falls from 29.50\% to 24.40\%, a change of $-5.10$\% ($n=1000$, $p=9\times10^{-4}$).

\paragraph{Interventions on the cost side.}Why the success rate of the frozen V-JEPA 2 falls while its readout improves is unresolved. Because the readout improves, the information is in the latent state, so we tested whether the cost was at fault, with two interventions on the cost side. Whitening divides each latent dimension by its standard deviation over the training data, so that a few high-variance dimensions do not dominate the distance. Reweighting measures the distance along the directions in which the linear head reads out the hand position, so that a difference in the hand affects the cost directly. Neither restored the success rate to the baseline level.

\section{Validity of the success criteria and resolution of the test}\label{app:metrics}
\paragraph{Smallest detectable difference.}All comparisons in this paper are paired tests on identical trial pairs (Section~\ref{sec:setup}). Such a test becomes significant when the two models being compared disagree on at least six trials and all six favor one model. With 2000 trial pairs this corresponds to a difference of 0.3\%, and with 1000 trial pairs to 0.6\%. This is the lower bound set by the test, not the precision of the measurement.
\paragraph{Validity of the success criteria.}We defined the success criteria of stack and coffee ourselves, so we check that the choice of tolerance does not determine the results. Table~\ref{tab:judge-validity} varies the tolerance of these two criteria and reports the success rate of the baseline model and the effect of each auxiliary-loss model. During evaluation, a trial stops when the environment judges it a success, and its state at that moment is saved. Re-judging under a different tolerance re-judges these saved states; it does not re-plan toward that tolerance. We check two points. First, the improvement does not depend on the tolerance. The models that improved significantly in the main text are the model supervised with both quantities in stack and the model supervised with the object in coffee (Table~\ref{tab:official}). Both improve significantly at every object tolerance from 4\,cm down to 1\,cm. When the hand tolerance is loosened to 3\,cm and 4\,cm and the baseline success rate rises to 48\% and 59\% in stack and to 49\% and 60\% in coffee, the improvement of the stack model remains significant, and that of the coffee model is significant up to 3\,cm. When the hand tolerance is tightened to 1\,cm, both models fail in almost every trial: the baseline succeeds in 0.7\% of the trials in stack and 4.7\% in coffee, so an effect of the auxiliary loss, if any, cannot appear in the success rate. Second, the tolerance is not too loose. A criterion that is too loose counts states unrelated to the goal as successes. We check this with two rates. The first is the fraction of trials that already satisfy the criterion at the start: with a hand tolerance of 2\,cm it is 1.7\% in stack and 9.0\% in coffee, against 5.4\% and 22.4\% with 4\,cm. The second is the false-match rate, the fraction of states from other episodes that are judged a success for a given goal. We measured it as follows: for each of the 2000 goals, we drew 200 states from other episodes of the dataset, applied the criterion to each as if it were the final state of a trial toward this goal, and counted the states judged a success. The false-match rate is 0.15\% in stack and 4.7\% in coffee. It is higher in coffee because the pod condition alone accepts 24.6\% of the states from other episodes. This is presumably because the pod ends inserted in the machine in every episode, so that its position is nearly the same across episodes and does not identify the goal. Adding the hand condition lowers the false-match rate to 4.7\%, so in coffee it is the hand condition that identifies the goal state, and the hand tolerance of 2\,cm cannot be loosened.

\begin{table}[h]\centering\small
\caption{\textbf{Validity of the success criteria}. Baseline success rate and effect of the auxiliary-loss models as the tolerance of the stack and coffee criteria is varied (identical 2000 trial pairs; $^{*}$: $p<0.05$), and the false-match rate, the fraction of states from other episodes that are judged a success, shown only for the tolerance used in the main text ($\dagger$). The auxiliary-loss models are those of Table~\ref{tab:official}.}
\label{tab:judge-validity}
\setlength{\tabcolsep}{4pt}
\begin{tabular}{llrrrrr}
\toprule
Task & Tolerance of the criterion & Baseline [\%] & \multicolumn{3}{c}{$\Delta$ of auxiliary-loss models [\%]} & False match [\%] \\
\cmidrule(lr){4-6}
\midrule
\multicolumn{3}{l}{\textit{stack (object $\land$ hand)}} & both & hand only & object only & \\
& object 1\,cm (hand 2\,cm) & 16.6 & $+2.45^{*}$ & $-0.05$ & $-3.95^{*}$ & --- \\
& object 2\,cm (hand 2\,cm) & 24.3 & $+4.20^{*}$ & $-1.60$ & $-1.45$ & --- \\
& object 3\,cm (hand 2\,cm) & 28.1 & $+4.75^{*}$ & $-1.90$ & $-0.60$ & --- \\
& object 4\,cm (hand 2\,cm)$\dagger$ & 30.1 & $+5.05^{*}$ & $-1.30$ & $-0.40$ & 0.15 \\
& hand 1\,cm (object 4\,cm) & 0.7 & $+0.15$ & $+0.10$ & $+0.40^{*}$ & --- \\
& hand 3\,cm (object 4\,cm) & 48.3 & $+4.80^{*}$ & $-3.95^{*}$ & $-0.30$ & --- \\
& hand 4\,cm (object 4\,cm) & 58.7 & $+3.15^{*}$ & $-5.30^{*}$ & $+0.50$ & --- \\
\midrule
\multicolumn{3}{l}{\textit{coffee (pod $\land$ hand)}} & both & object only & & \\
& pod 1\,cm (hand 2\,cm) & 16.2 & $+0.95$ & $+2.40^{*}$ & & --- \\
& pod 2\,cm (hand 2\,cm) & 26.2 & $+0.50$ & $+4.05^{*}$ & & --- \\
& pod 3\,cm (hand 2\,cm) & 31.4 & $-0.20$ & $+4.00^{*}$ & & --- \\
& pod 4\,cm (hand 2\,cm)$\dagger$ & 34.1 & $-0.60$ & $+3.90^{*}$ & & 4.7 \\
& hand 1\,cm (pod 4\,cm) & 4.7 & $-0.50$ & $-0.15$ & & --- \\
& hand 3\,cm (pod 4\,cm) & 49.0 & $\pm0.00$ & $+2.60^{*}$ & & --- \\
& hand 4\,cm (pod 4\,cm) & 60.2 & $+0.90$ & $+1.30$ & & --- \\
\bottomrule
\end{tabular}
\end{table}

\begin{figure}[h]
\centering
\includegraphics[width=\linewidth]{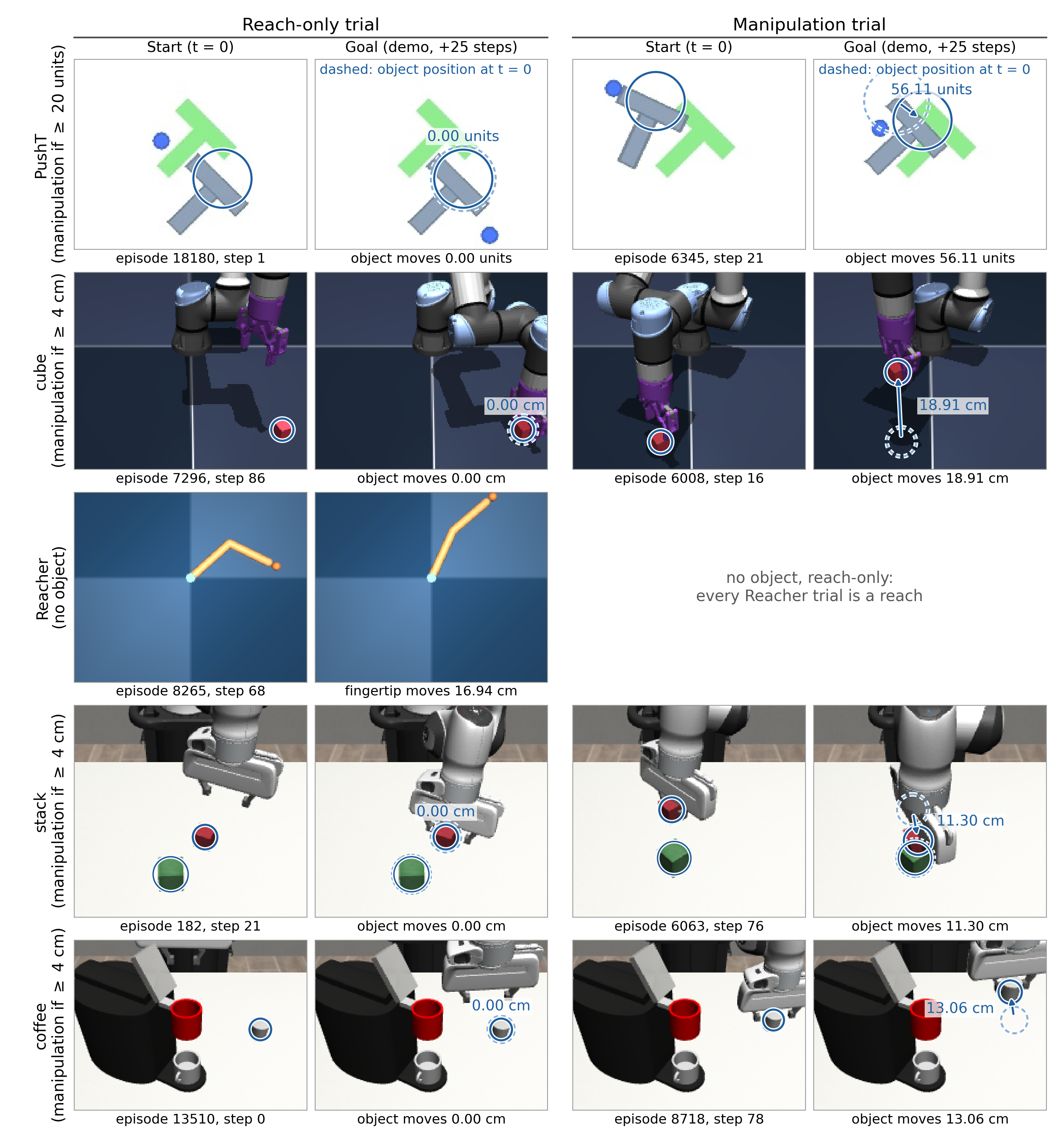}
\caption{\textbf{Examples of the trial strata (five tasks)}. For each task, the start frame and the goal frame of the demonstration for a reach-only trial and a manipulation trial. The strata are those of Table~\ref{tab:dissociation}; a trial whose object displacement is at least the object tolerance (4\,cm for stack, coffee, and cube; 20\,units for PushT) is a manipulation trial. Solid circles mark the object position and dashed circles the position at the start; the number next to the arrow is the displacement from start to goal. In reach-only trials the dashed circle coincides with the solid one. The circle in PushT marks the object position, not the outline of the T. Reacher has no object, all of its trials are reach-only, and no marks are drawn. Because the criterion of cube has no hand term, its reach-only trials satisfy the criterion at the start.}
\label{fig:strata-all}
\end{figure}

\paragraph{Examples of the trial strata.}Figure~\ref{fig:strata-all} shows, for each of the five tasks, the start frame and the goal frame of the demonstration for a reach-only trial and for a manipulation trial. The stratum is decided by whether the object position at the start lies within the object tolerance of the criterion of its position at the goal (Section~\ref{sec:dissociation}). Reacher has no object, so all of its trials are reach-only; the criterion of cube has no hand term, so its reach-only trials already satisfy the criterion at the start.

\section{Supporting experiments}\label{app:extra}

\paragraph{Readout of the supervised and the unsupervised quantities.}Section~\ref{sec:dissociation} showed the selectivity of the effect of the auxiliary loss in the success rate; here we show the same selectivity in the readout. Supervising only some of the success-criterion quantities also changes the readout of the quantities that were not supervised. Table~\ref{tab:crossreadout} lists, for the three stack models of Table~\ref{tab:dissociation}, namely the baseline, the hand-only model ($\lambda=1$), and the object-only model ($\lambda=0.1$), the readout errors of the hand and of the two cubes from the encoder output. A quantity that moves together with the supervised quantity becomes more readable than in the baseline: supervising the hand improves the readout of the cube that the hand carries, and supervising the objects improves the readout of the hand that carries them. The one exception is the base cube, which the hand never touches: supervising only the hand raises its readout error from 0.96\,cm to 3.49\,cm, and the share of frames beyond the 4\,cm tolerance from 3\% to 43\%. The direction is the same for the terminal latent state that the planner scores: with the object supervised, the hand readout error there falls from 1.47\,cm to 1.13\,cm, and the share beyond the 2\,cm tolerance from 26\% to 15\% ($n=1600$). The degradation of the reach-only trials under object supervision (Section~\ref{sec:dissociation}) is therefore not caused by the hand becoming less readable.

\begin{table}[h]
\centering
\caption{\textbf{Readout error from the latent state (stack)}. 5000 frames not used for training, linear ridge regression, median of the 3D distance [cm]. Bold marks the supervised quantity. Tolerances are 2\,cm for the hand and 4\,cm for the cubes. The error of a constant prediction is 8.93 for the hand, 7.56 for the carried cube, and 6.96\,cm for the base cube.}
\label{tab:crossreadout}
\small
\begin{tabular}{lccc}
\toprule
Model & Hand & Carried cube & Base cube \\
\midrule
baseline & 3.15 & 1.82 & 0.96 \\
supervised by the hand only ($\lambda=1$) & \textbf{0.14} & 1.05 & 3.49 \\
supervised by the object only ($\lambda=0.1$) & 1.96 & \textbf{0.29} & \textbf{0.18} \\
\bottomrule
\end{tabular}
\end{table}

\begin{figure}[h]\centering
\includegraphics[width=\linewidth]{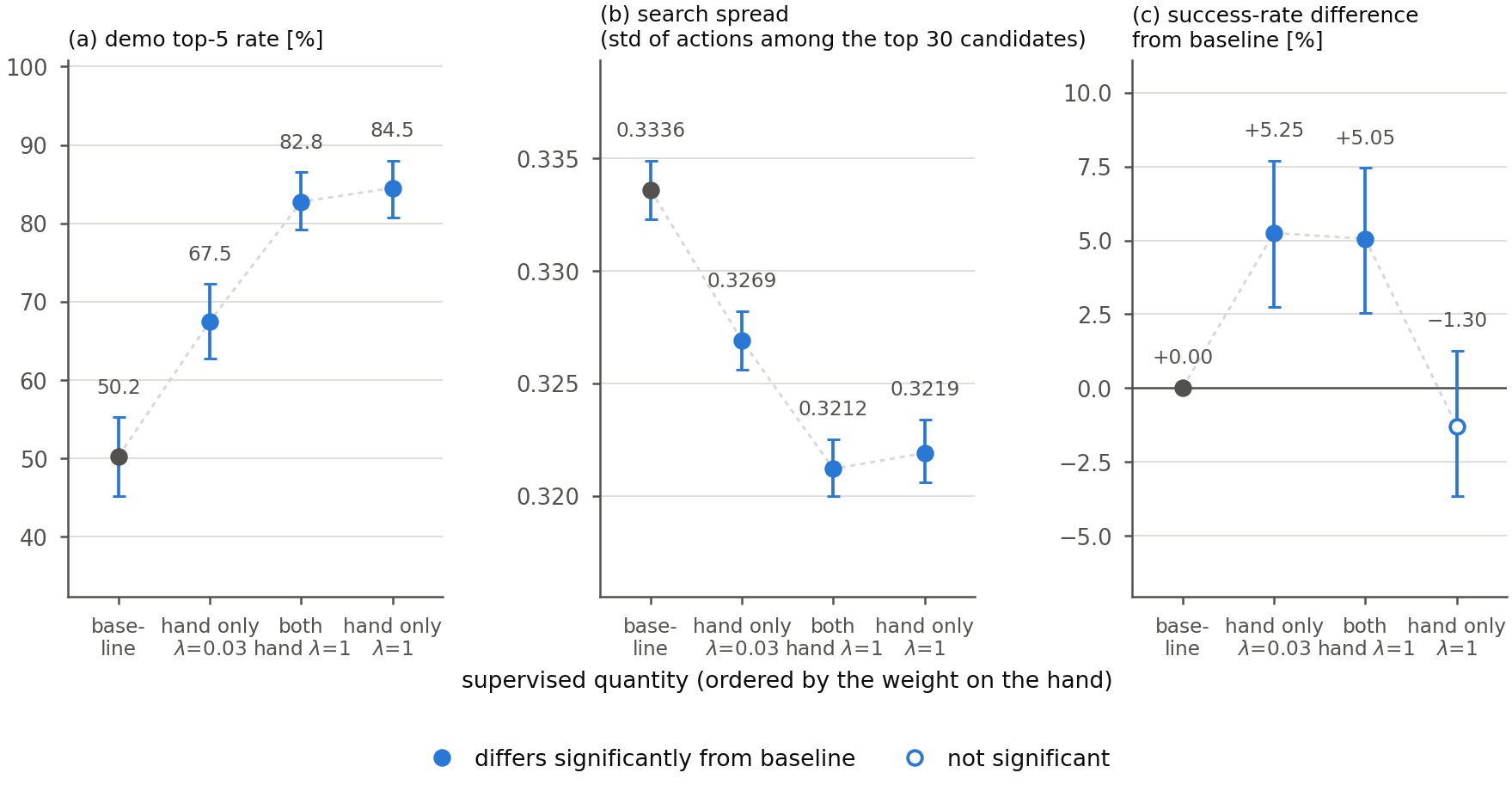}
\caption{\textbf{Ranking of candidates, shape of the search, and success rate (stack)}. The horizontal axis orders the baseline and three auxiliary-loss models by the weight on the hand (the model supervised with both quantities has object $\lambda=0.1$). (a) Demo top-5 rate: the fraction of trials in which the demonstration action sequence ranks in the top five by cost among itself and 63 noisy copies (400 trials). (b) Search spread: the standard deviation of the actions among the top 30 candidates of the final CEM iteration in the actual planning; the smaller, the more focused the search (first plan of 2000 trials). (c) Difference in success rate from the baseline [\%] (2000 trials). Intervals are 95\%; filled markers mark a difference from the baseline whose interval excludes 0 (in (c), two-sided McNemar $p<0.05$). (a) and (b) follow the weight on the hand; (c) does not.}\label{fig:ranking}\end{figure}

\paragraph{Ranking of candidates and shape of the search.}The auxiliary loss improves the ranking of candidate action sequences by the planning cost, but the size of this improvement does not follow the size of the improvement in success rate (Figure~\ref{fig:ranking}; the basis of the third open problem in Section~\ref{sec:limitations}). We measure the quality of the ranking as follows. Taking the action sequence of the demonstration of each trial and 63 noisy copies of it as candidates, we call the fraction of trials in which the demonstration ranks in the top five by cost the demo top-5 rate. We measure the shape of the search by the standard deviation of the actions among the top 30 candidates of the final CEM iteration in the actual planning, and call it the search spread; the smaller, the more focused the search. The demo top-5 rate in Figure~\ref{fig:ranking}(a) is 50\% for the baseline, 68\% for the hand-only model with $\lambda=0.03$, 83\% for the model supervised with both quantities (hand $\lambda=1$, object $\lambda=0.1$), and 85\% for the hand-only model with $\lambda=1$, rising with the weight on the hand (400 trials). The search spread in Figure~\ref{fig:ranking}(b) also shrinks as the weight on the hand grows (first plan of each of 2000 trials per model). The success-rate differences in Figure~\ref{fig:ranking}(c), however, are $+5.25$\%, $+5.05$\%, and $-1.30$\%: the hand-only model with $\lambda=1$, whose ranking and search are the best, is the lowest. Hence the changes in ranking and in the shape of the search do not identify the route by which the auxiliary loss raises the success rate.

\paragraph{Component decomposition of PushT.}The official criterion of PushT is decided almost entirely by the agent position. The criterion combines the agent position and the block position into one norm and thresholds it at 20 units, so the criterion itself cannot tell which component improved. Table~\ref{tab:pusht} re-aggregates the same 2000 trial pairs by component. The improvements on PushT stated in the main text are the numbers in the official-criterion column of Table~\ref{tab:pusht}; the component columns serve only to show which component changed, and their thresholds (agent $<10$; block $<15$ and angle $<20^{\circ}$) were set by us from a threshold sweep. The model supervised with the hand improves only the agent component significantly, and the block component, which it does not supervise, changes by $+0.1$\%. The model supervised with the object does not move the official criterion. The reason lies in the criterion itself: decomposing the squared norm into its components, the agent's share is 86\% at the median over the baseline's 2000 trials, and the agent's term exceeds the block's in 83\% of the trials. Moreover, the readout error of the block position from the baseline, 10.8\,units, lies inside the component threshold of 15, so the block is already readable at the precision of that threshold without supervision.

\begin{table}[h]
\centering
\caption{\textbf{Success rate of PushT by component of the criterion}. Identical 2000 trial pairs. The improvements stated in the main text are the numbers in the official-criterion column; the component columns show which component changed. We set the component thresholds to agent $<10$, block $<15$, and angle $<20^{\circ}$.}
\label{tab:pusht}
\begin{tabular}{lccc}
\toprule
Supervised quantity & Official criterion & Agent component & Block component \\
\midrule
baseline & 91.5\% & 23.5\% & 93.3\% \\
hand only & $93.5\,(+2.0^{*})$ & $\mathbf{28.5\,(+5.0^{*})}$ & $93.4\,(+0.1)$ \\
object only & $91.6\,(+0.2)$ & $24.1\,(+0.6)$ & $93.6\,(+0.3)$ \\
both & $\mathbf{95.0\,(+3.5^{*})}$ & $26.1\,(+2.6^{*})$ & $94.8\,(+1.5^{*})$ \\
\bottomrule
\end{tabular}
\end{table}

\paragraph{End states of the degraded trials.}The complementary degradation of Section~\ref{sec:dissociation} appears in the state at the end of the trial: the supervised quantity ends closer to the goal, and the unsupervised quantity ends farther from it than in the baseline (Table~\ref{tab:failterm}). We compare the models on the trials that both fail, because including the trials the baseline succeeds in would make its errors small by definition. On the manipulation trials, the hand-only model ends with the hand closer to the goal than the baseline (3.19\,cm against 5.07\,cm) but the object farther (11.07\,cm against 9.83\,cm), and the object enters its tolerance somewhere along the path in fewer trials (15\% against 27\%). On the reach-only trials, where the object is inside its tolerance from the start, the object-only model ends with the hand farther from the goal than the baseline (5.05\,cm against 4.36\,cm).

\begin{table}[h]
\centering
\caption{\textbf{End states of the trials that both models fail (stack)}. Only the two combinations in which the complementary degradation occurs are shown: the hand-only model on the manipulation trials, and the object-only model on the reach-only trials. The end state is at step 50. Errors are medians of the 3D distance [cm]. Tolerances are 2\,cm for the hand and 4\,cm for the object, where the object error is the larger error of the two cubes. ``Inside tolerance'' is the fraction of trials in which the quantity entered its tolerance at some point along the path. Strata are as in Table~\ref{tab:dissociation}.}
\label{tab:failterm}
\small
\begin{tabular}{llccccc}
\toprule
Stratum & Model & $n$ & Hand error & Object error & Hand inside tolerance & Object inside tolerance \\
\midrule
Manipulation & baseline & 794 & 5.07 & 9.83 & 20.8\% & 26.7\% \\
 & hand only & 794 & 3.19 & 11.07 & 37.0\% & 14.9\% \\
\midrule
Reach-only & baseline & 382 & 4.36 & 0.09 & 0.8\% & 100\% \\
 & object only & 382 & 5.05 & 0.04 & 0.0\% & 100\% \\
\bottomrule
\end{tabular}
\end{table}

\paragraph{Failure types in coffee and stack.}Supervising the hand does not move the success rate on coffee because the failures of the baseline on the reach-only trials are of a different type than on stack. We divide the failures into three types: the hand approaches within 4\,cm of the goal along the path and stops short there; the hand approaches and then ends at least 2\,cm farther than its closest point; and the hand never approaches within 4\,cm. On stack the failures are 57\%, 15\%, and 28\% in this order, so stopping short dominates. On coffee they are 12\%, 44\%, and 44\%, so departing and never approaching dominate. What supervising the hand repairs is the inability of the cost to discriminate the hand position finely. The failures that this can turn into successes are those that stop short; failures that depart or never approach are not repaired by a finer readout of the hand. On coffee these two types make up 88\%. The median hand error of the approaching failures over time shows the difference: on stack it keeps approaching slowly, 4.2\,cm at step 25, the end of the first plan, and 3.6\,cm at step 50, whereas on coffee it reaches its minimum of 3.6\,cm at step 25 and then, after replanning, departs to 8.4\,cm at step 50. Indeed, the hand-only model has nearly the same composition of failures: 15\% stop short, 50\% depart, and 34\% never approach. That the departing failures do not decrease also shows in the gap between the world model's prediction and the actual end state. For the action sequence chosen by the planner, the hand read out from the terminal state predicted by the world model is 2.9\,cm from the goal for the baseline and 2.1\,cm for the hand-only model, whereas the hand at the executed end state is 6.3\,cm and 4.9\,cm away; the gap between prediction and execution remains about 3\,cm with or without supervising the hand (200 reach-only trials, of which the prediction could be reconstructed for 133). Supervising the hand brings the hand at the end state 1.4\,cm closer to the goal, but this is far from the 2\,cm tolerance and does not turn the trial into a success. What the auxiliary loss can turn into successes are failures that stop just outside the tolerance, and coffee has few of them. Supervising the hand only makes the hand readable from the world model's prediction; it does not make the execution follow the prediction.

\begin{figure}[h]\centering
\includegraphics[width=\linewidth]{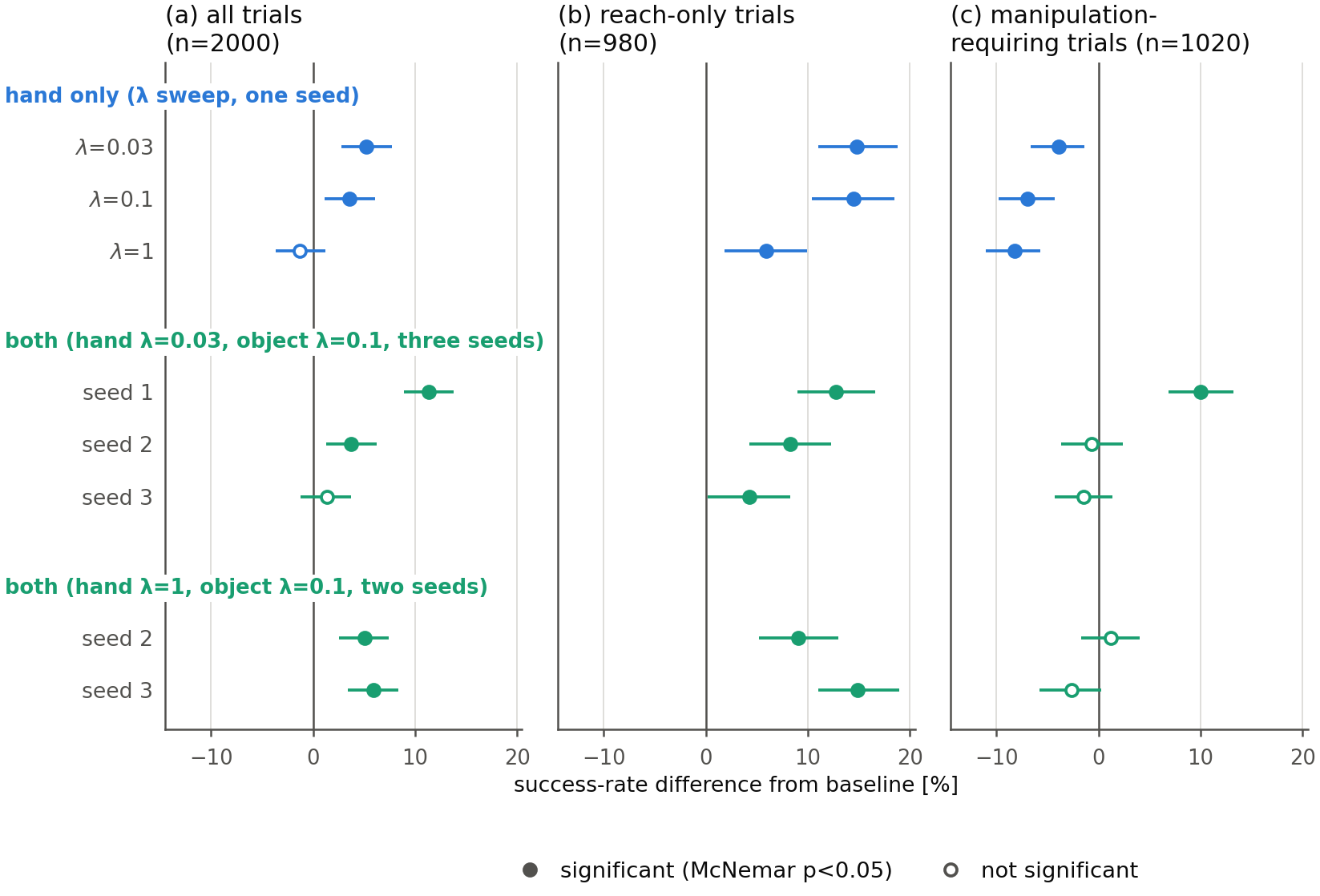}
\caption{\textbf{Effect by trial stratum for eight auxiliary-loss models on stack}. Difference in success rate from the baseline [\%] on identical 2000 trial pairs; the strata have 980 reach-only and 1020 manipulation trials. Includes the hand-only ($\lambda=1$) and the both ($\lambda=1$, seed 2) models of Table~\ref{tab:dissociation}. Seeds 1--3 are three training seeds, the same three for every arm. Bars are 95\% intervals from a paired bootstrap over trials; filled markers are significant by two-sided paired McNemar ($p<0.05$). The criterion is as in Table~\ref{tab:dissociation}.}\label{fig:stack12}\end{figure}

\paragraph{Effects by trial stratum for the remaining models.}The complementary degradation under supervision of the hand alone, and the absence of degradation under supervision of both quantities, hold across the weight and the training seed (Figure~\ref{fig:stack12}; eight stack models). The three hand-only models, trained with the same seed and $\lambda=0.03$, $0.1$, and $1$, all improve significantly on the reach-only trials and all degrade significantly on the manipulation trials. However, as $\lambda$ grows, the improvement on the reach-only trials shrinks ($+14.8$, $+14.5$, $+5.9$\%) and the degradation on the manipulation trials grows ($-3.9$, $-7.0$, $-8.2$\%), and at $\lambda=1$ the overall improvement disappears. The five models supervised with both quantities, three seeds at hand $\lambda=0.03$ and two seeds at hand $\lambda=1$ (object $\lambda=0.1$), all improve significantly on the reach-only trials and none degrades significantly on the manipulation trials. The direction of the effect does not depend on the seed, but its size does: the overall improvements of the three seeds at $\lambda=0.03$ are $+11.4$, $+3.7$, and $+1.4$\%.

\begin{figure}[h]\centering
\includegraphics[width=\linewidth]{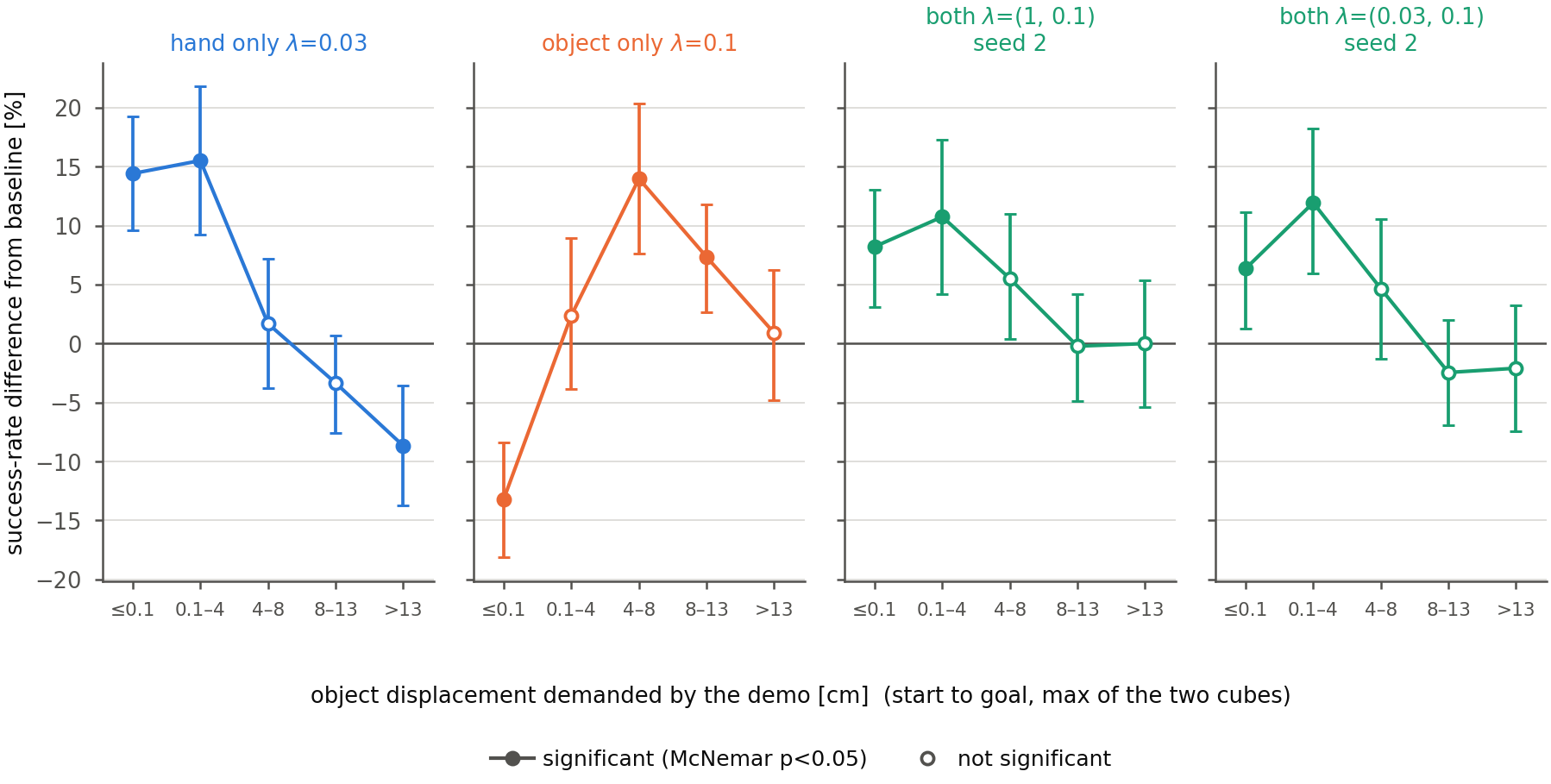}
\caption{\textbf{Demanded object displacement and the effect of the auxiliary loss} (stack). Four auxiliary-loss models on stack, namely hand only with $\lambda=0.03$, object only with $\lambda=0.1$, both with $\lambda=(1,0.1)$, and both with $\lambda=(0.03,0.1)$ and seed 2, evaluated on the same 2000 trials binned by displacement. The horizontal axis is the bin of the start-to-goal displacement of the object, the larger of the two cubes. The vertical axis is the difference in success rate from the baseline [\%]; bars are 95\% intervals from a paired bootstrap over trials, and filled markers are significant by two-sided paired McNemar ($p<0.05$). The five bins contain 645, 335, 236, 449, and 335 trials.}\label{fig:dose}\end{figure}

\paragraph{Demanded object displacement and the effect.}The effect of the auxiliary loss concentrates on the trials in which the supervised quantity is tested. We divided the 2000 stack trials by the start-to-goal displacement of the object into five bins, at most 0.1\,cm, 0.1--4, 4--8, 8--13, and above 13\,cm, and measured the effect in each bin (Figure~\ref{fig:dose}). The first two bins make up the reach-only trials of Table~\ref{tab:dissociation} and the last three the manipulation trials. The effect of the hand-only model is largest in the 0.1--4\,cm bin, decreases monotonically as the displacement grows, and becomes a significant degradation above 13\,cm. The model supervised only with the object shows the mirror image: a significant degradation in the bin at most 0.1\,cm and significant improvements in the 4--13\,cm bins. The two models supervised with both quantities improve in the bins of small displacement and do not degrade in the bins of large displacement. That the improvement concentrates on trials demanding little motion is consistent with the observation on coffee (the paragraph on failure types in coffee and stack) that the auxiliary loss can only turn into successes the failures that stop close to the goal.

\section{Item-by-item comparison with the closest work}\label{app:positioning}
Table~\ref{tab:positioning} compares this work with the three closest works discussed in Section~\ref{sec:related} item by item. The items are whether the method plans with a latent world model, whether it adds an auxiliary loss, what the loss supervises, whether the supervised target is fixed by the success criterion, whether the loss is applied to the predicted latent state, whether inference is unchanged, and whether the effect is measured per supervised quantity and per trial stratum. Only this work restricts the target to the quantities in the success criterion and measures the effect per quantity and per stratum.

\begin{table}[!htbp]\centering
\caption{\textbf{Position relative to the closest work}. ``Auxiliary loss'' is a supervised loss added to the main objective of latent prediction; the rollout loss of V-JEPA 2-AC is not counted because its target is the latent state itself. ``Applied to predicted latent'' is whether the auxiliary loss shapes the latent state output by the predictor, which is what the planning cost scores: PhyLatent and this work apply the loss to both the encoder and the predictor outputs, PSG-JEPA only to the encoder output, and RC-aux stops the gradient on the predictor side. ``Per quantity and stratum'' is whether the effect is measured separately for each supervised quantity and for each trial stratum. ``---'' means the design element is absent. PhyLatent and PSG-JEPA are concurrent work released in August 2026. The planner of PSG-JEPA is an amortized inverse-dynamics model that uses neither the predictor nor the latent-distance cost, so it is not counted as planning with a latent world model.}
\label{tab:positioning}
\scriptsize
\setlength{\tabcolsep}{3.5pt}
\renewcommand{\arraystretch}{1.15}
\resizebox{\linewidth}{!}{%
\begin{tabular}{lccccccc}
\toprule
& \shortstack{Plans with\\latent WM} & \shortstack{Auxiliary\\loss} & Supervised target & \shortstack{Target fixed by\\the success criterion} & \shortstack{Applied to\\predicted latent} & \shortstack{Inference\\unchanged} & \shortstack{Per quantity\\and stratum} \\
\midrule
LeWM, DINO-WM, V-JEPA 2-AC & \cmark & \xmark & --- & --- & --- & --- & --- \\
RC-aux & \cmark & \cmark & reachability & \xmark & \xmark & \xmark & \xmark \\
PhyLatent & \cmark & \cmark & full simulator state & \xmark & \cmark & \cmark & \xmark \\
PSG-JEPA & \xmark & \cmark & proprioceptive state & \xmark & \xmark & \cmark & \xmark \\
\textbf{Ours} & \cmark & \cmark & \textbf{only success-criterion quantities} & \cmark & \cmark & \cmark & \cmark \\
\bottomrule
\end{tabular}}
\end{table}

\end{document}